\documentclass[letterpaper]{article} 
\usepackage{aaai25}  
\usepackage{times}  
\usepackage{helvet}  
\usepackage{courier}  
\usepackage[hyphens]{url}  
\usepackage{graphicx} 
\usepackage{natbib}  
\usepackage{caption} 
\usepackage{algorithmic}

\usepackage[ruled,vlined,noresetcount]{algorithm2e}
\usepackage{amsfonts}       
\usepackage{amsmath}        
\usepackage{bm}             
\usepackage{booktabs}       
\usepackage{dsfont}         
\usepackage[utf8]{inputenc} 
\usepackage{microtype}      
\usepackage{multirow}       
\usepackage{nicefrac}       
\usepackage{paralist}       
\usepackage[table]{xcolor}  

\newtheorem{definition}{Definition} 

\SetKwComment{Comment}{// }{}
\RestyleAlgo{ruled}
\usepackage{newfloat}
\usepackage{listings}

\def\ie{{\em i.e.}\ } 
\def\eg{{\em e.g.}\ }

\definecolor{darker}{rgb}{0.357,0.608,0.835}
\definecolor{dark}{rgb}{0.824,0.87,0.937}
\definecolor{light}{rgb}{0.917,0.937,0.968}
\definecolor{subtotal}{rgb}{0.76, 0.8, 0.88}
\definecolor{total}{rgb}{0.6, 0.75, 0.88}

\DeclareCaptionStyle{ruled}{labelfont=normalfont,labelsep=colon,strut=off}

\title{FedGA: Federated Learning with Gradient Alignment for Error Asymmetry Mitigation}
\author{
    Chenguang Xiao, Zheming Zuo\correspond, Shuo Wang
}
\affiliations {
    School of Computer Science, University of Birmingham, UK\\
    cxx075@student.bham.ac.uk, \{z.zuo.1, s.wang.2\}@bham.ac.uk
}

\begin{document}

\maketitle

\begin{abstract}

    Federated learning (FL) triggers intra-client and inter-client class imbalance, with the latter compared to the former leading to biased client updates and thus deteriorating the distributed models. Such a bias is exacerbated during the server aggregation phase and has yet to be effectively addressed by conventional re-balancing methods.  To this end, different from the off-the-shelf label or loss-based approaches, we propose a gradient alignment (GA)-informed FL method, dubbed as FedGA, where the importance of error asymmetry (EA) in bias is observed and its linkage to the gradient of the loss to raw logits is explored. Concretely, GA, implemented by label calibration during the model backpropagation process, prevents catastrophic forgetting of rate and missing classes, hence boosting model convergence and accuracy. Experimental results on five benchmark datasets demonstrate that GA outperforms the pioneering counterpart FedAvg and its four variants in minimizing EA and updating bias, and accordingly yielding higher F1 score and accuracy margins when the Dirichlet distribution sampling factor $\alpha$ increases. The code and more details are available at \url{https://anonymous.4open.science/r/FedGA-B052/README.md}.
\end{abstract}

%

\section*{Introduction}
\label{sec:intro}



Federated learning (FL) utilizes data stored on edge devices to enable knowledge extraction \cite{zhang2024upload} from distributed data while maintaining privacy \cite{Zuo2021DataAnon}. However, FL faces significant challenges due to data heterogeneity \cite{shi2024clip,wang2024dfrd}, particularly class imbalance \cite{wu2024solving}, where classes are unequally represented in the training data. Class imbalance issues are prevalent in various real-world applications, including online grooming detection \cite{zuo19AdaTSK}, clinical diagnosis \cite{wu2023fediic}, brand anti-counterfeiting \cite{Zuo_DCF_2024_CVPRW} etc. Noteworthy, FL deployed in distributed agents tend to enhance model performance with privacy preserved \cite{10574838}, yet class imbalance issues can hinder minority class prediction precision and slow model convergence \cite{wang2024turbosvm}.



\begin{figure}[htbp]
    \centering
    \includegraphics[width=0.96\linewidth]{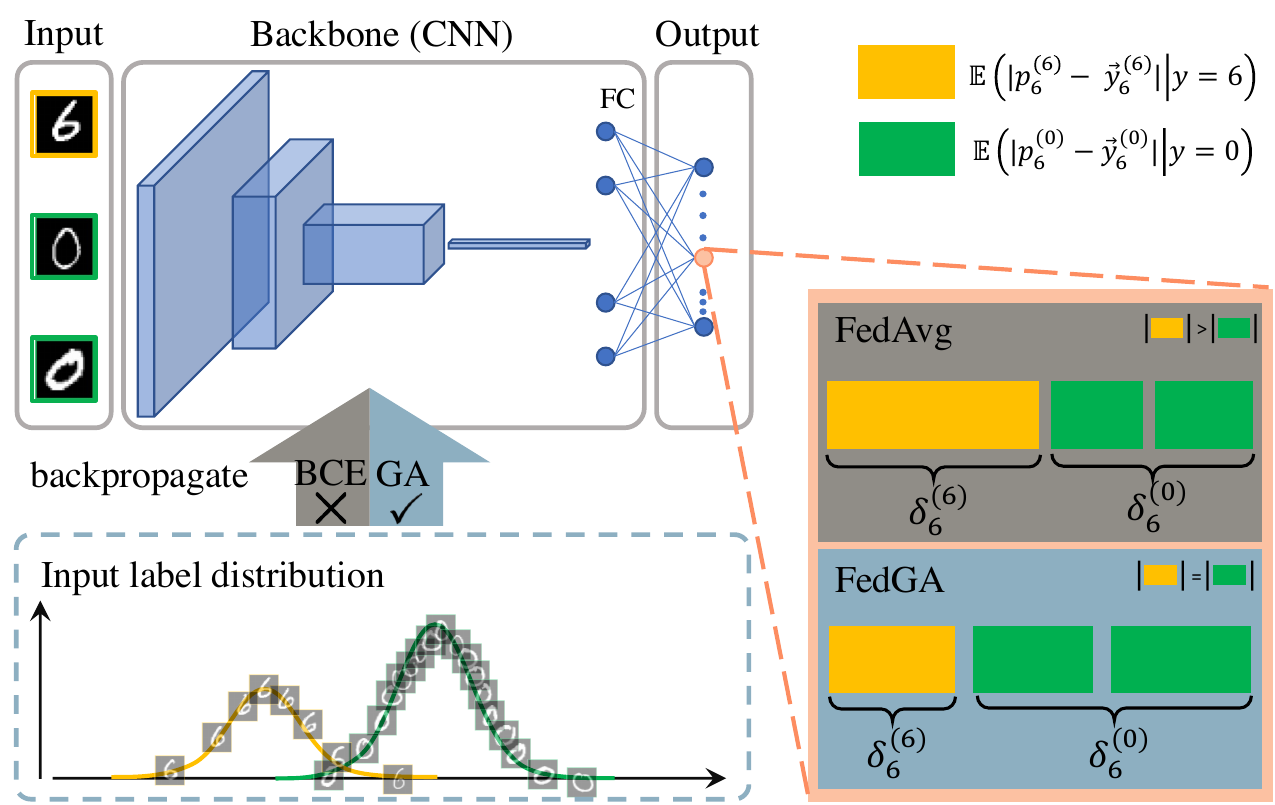}
    \caption{
        We propose an FL framework for error asymmetry mitigation caused by class imbalance. Our gradient alignment (GA) method, different from binary cross entropy (BCE) adopted in the prestigious FedAvg, scales active and inactive gradients through the label calibration to guide the model backpropagation process, thus reducing client update bias and allaying privacy and communication concerns.
    }
    \label{fig:mot}
\end{figure}

Class imbalance in FL typically manifests as intra-client and inter-client imbalances \cite{shen2021agnostic,wang2021addressing,10492865}. The former analogizes the traditional class imbalance in centralized learning (CL), and the latter refers to mismatched class distributions between clients, \eg one client having only class 0 data and another having only class 1 data, adding complexity to the issue. The investigation of inter-client class imbalance is nascent. Conventional re-balancing techniques, \eg oversampling and undersampling from CL, are often ineffective due to system constraints \cite{xiao2023triplets}, particularly in addressing missing classes. Besides, latest works \cite{sarkar2020fed,yang2021federated,shuai2022balancefl,zhang2022federated} mitigated the class imbalance, most of which modified the loss for specific classes or clients to rebalance intra- or inter-client imbalance. However, these inter-client re-balancing approaches involve potential privacy risks and demand additional communication and computation costs \cite{xiao2024fed}.

Inter-client imbalance on the server causes biased client updates, distorting its aggregation. For instance, in FedAvg \cite{mcmahan2017communication}, averaging biased updates results in poor global gradient estimation \cite{su2023non}, slowing convergence and reducing accuracy. On the client side, catastrophic forgetting \cite{babakniya2024data} of rare or missing classes further contributes to biased updates. Furthermore, the FL system requires minimal information exchange, constrained communication, and flexible participation, which filter out or invalidate most CL and FL methods. For example, the missing class problem undermines the effectiveness of re-sampling and re-weighting in the FL system. Even modified loss functions fail to re-balance the client and reduce the update bias as the missing class is not present in the client dataset. From there, as depicted in Fig. \ref{fig:mot}, this work proposes a gradient alignment approach to cope with the aforementioned challenge, and the main contributions are three-fold:

\begin{itemize}
    \item We prove that class imbalance in FL is fundamentally associated with asymmetric Type I and Type II errors as the gradient of the raw logits is composed of mismatched active and inactive ones corresponding to these errors.
    \item We propose a gradient alignment (GA) algorithm via label calibration to scale the active and inactive gradients, named FedGA, to eliminate error asymmetry (EA) as the first class-wise bias reduction algorithm free of extra privacy and communication concerns in FL.
    \item We show that FedGA is superior to five state-of-the-art algorithms in reducing EA and client update bias while improving convergence speed, accuracy, and F1 score over five benchmark datasets with a high degree of self-explainability and scalability for decentralized AI.
\end{itemize}


\section*{Related Works} \label{sec:related_works}


The class imbalance in FL shares similarities with CL but presents distinct challenges. While several studies have addressed that in CL \cite{elkan2001foundations,chawla2002smote,he2008adasyn,liu2008exploratory}, fewer have focused on FL, which are categorized into data- and model-level methods.


\noindent \textbf{Data-Level Approaches.} $\quad$ This category of approaches addresses sample imbalance mostly through random oversampling of the minority class or undersampling of the majority class~\cite{liu2008exploratory}. Heuristic re-sampling methods~\cite{tang2015enn} target difficult minority examples, while the synthetic minority oversampling technique (SMOTE) \cite{chawla2002smote} and its derivatives generate synthetic minority samples to mitigate information loss and overfitting \cite{he2008adasyn}. In addition, generative adversarial network (GAN)-based methods~\cite{ali2019mfc} were employed to produce synthetic minority samples.


\noindent \textbf{Model-Level Approaches.} $\quad$ Unlike data-level approaches, model-level approaches address the issue during training. Simple adjustments, \eg altering class weights or modifying the decision threshold, have been proven to mitigate the issue \cite{elkan2001foundations}. The use of specialized loss functions for class imbalance, \eg focal loss~\cite{lin2017focal} and class-balanced loss~\cite{cui2019class}, further improves performance under label skew. Besides, ensemble learning~\cite{feng2018class}, one-class learning \cite{lee2023resampling}, and transfer learning \cite{li2024gated} were promising solutions.

\noindent \textbf{FL-Specific Approaches.} $\quad$ At present, limited researches address this issue in FL, focusing on intra-client (local) and inter-client (global) imbalances. Several studies further explore inter-client imbalance as the mismatch between local and global imbalances~\cite{wang2021addressing,xiao2021experimental}. Loss function-based approaches are commonly adopted in FL. Fed-focal koss \cite{sarkar2020fed} combines focal loss with client selection to improve minority class performance, while ratio loss (FedRL) \cite{wang2021addressing} addresses global imbalance by introducing a ratio parameter to indicate imbalance levels, raising privacy concerns. \citeauthor{zhang2022federated} utilizes a modified \texttt{softmax} function for intra-client imbalance \cite{zhang2022federated} but does not fully cover the class-missing problem. Client selection methods \cite{yang2021federated}, such as Astraea \cite{duan2020self}, balance clients into groups, but add complexity and privacy risks. BalanceFL \cite{shuai2022balancefl}, a combination of knowledge distillation and other methods, addresses inter-client imbalance and missing classes, however, is not sufficiently generalizable. \citeauthor{dai2023tackling} tackles data heterogeneity through class prototypes \cite{dai2023tackling}, also raising privacy issues. CLIMB \cite{shen2021agnostic} leverages loss constraints to minimize client update bias by capturing the mismatch between local and global imbalances, but involves transmitting dual variables and local models to the server.

\section*{Asymmetric Error}
\label{sec:asymmetic_error}

\subsection{Types of Asymmetric Error}
\begin{figure}[htbp]
    \centering
    \includegraphics[width=\linewidth]{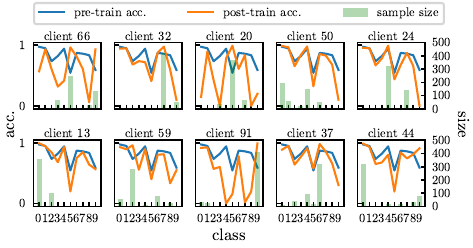}
    \caption{
        Class accuracy before and after local training with classical FedAvg on inter-client imbalanced MNIST. Each subplot corresponds to a random active client, showing sharp accuracy drops for rare and missing classes post-training.
    }
    \label{fig:pred_diverg}
\end{figure}

Inter-client imbalance is commonly deemed for producing biased updates from clients, thus compromising global model aggregation \cite{kim2024navigating}. Furthermore, the linear combination of biased updates results in an inaccurate estimation of the global gradient, thus hindering convergence \cite{varno2022adabest}. Fig.~\ref{fig:pred_diverg} illustrates the divergence of models trained on locally imbalanced data across various clients, revealing significant disparities in their effects on different classes. This results in catastrophic forgetting \cite{babakniya2024data}, particularly for minority and absent classes. Consequently, aggregating these divergent and biased client models poses a substantial challenge at the global level, as the weighted average operation used for model aggregation does not accurately reflect actual performance of the models.


Class imbalance leads to model bias for several reasons, notably inadequate sample size \cite{tao2024meanshift} and underestimated loss \cite{xiao2024cycle}. While data-level and algorithm-level approaches are employed to cope with this issue, label distribution and loss values do not directly measure model bias, making them insufficient for bias elimination. Instead of depending on input labels or class loss, we propose a direct measure of model bias, dubbed Error Asymmetry (EA), which is based on two types of errors. This measure informs a novel approach to effectively mitigate bias.


Typically, Type I error $u_{\text{I}}$ and Type II error $u_{\text{II}}$ refer to a false positive and a false negative, respectively. Practically, the error is commonly defined as the difference between the model prediction $p$ and the ground-truth label $y$. Hence, these two types of errors can be written as:
\begin{equation}
    \label{eq:type_error_1_n_2}
    \begin{cases}
        u_{\text{I}} = |p-1| & (y = 1), \\
        u_{\text{II}} = |p|  & (y=0).
    \end{cases}
\end{equation}

\subsection{Measurable Error Asymmetry}

Given an input data instance $x$, the corresponding groud-truth label $y$, and a binary classifier $f_\theta$, the prediction error $u$ is expressed by $|y - p|$ where $p$ equals to $f_{\theta}(x)$.
The prediction error of the minority class (\ie$u_{\text{I}}$) tends to be larger than that of the majority class (\ie$u_{\text{II}}$) since $\mathbb{E}(u|y=1) \gg \mathbb{E}(u|y=0)$. Indicating less confidence in the minority class, this aligns with the biased model achieving less competitive performance on the minority class.

\begin{figure}[htbp]
    \includegraphics[width=\linewidth]{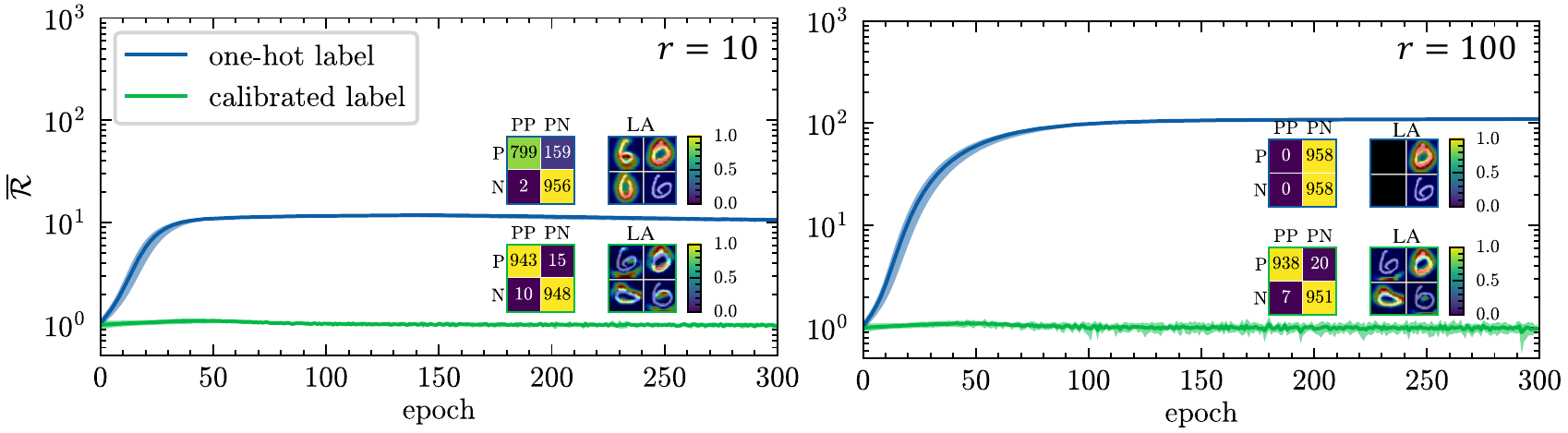}
    \caption{
        Mean error ratio between Type I to Type II error on an imbalanced binary (\eg class `6' and `0') subset of the MNIST dataset with imbalance ratio $r$ valued as 10 and 100. PP and PN represent predicted positive and negative, respectively. LA denotes layerwise attention. Noteworthy, FedGA (\textcolor{green}{green}) is equivalent to FedAvg (\textcolor{blue}{blue}) when $r$ equals 1. Best viewed in color and zoomed mode.}
    \label{fig:grad_ratio}
\end{figure}

Using an imbalanced MNIST~\cite{deng2012mnist} binary subset containing $N_0$ negative and $N_1$ positive samples, we logged two types of errors during training, as shown in Fig.~\ref{fig:grad_ratio}. Despite the imbalance ratios ($r$), the $u_{\text{I}}$ always converges to $r$ times that of $u_{\text{II}}$. As such, we term the ratio of Type I and II errors in class imbalance as EA $\mathcal{E}$ which is computed by $u_{\text{I}}/u_{\text{II}}$.

Apart from EA, the cumulative Type I and Type II errors always converge to 1, as the ratio of negative and positive samples is $r$. Equivalently, \( N_1 \mathbb{E}(e|y=1) \approx N_0 \mathbb{E}(e|y=0) \) holds for any \( r = \frac{N_0}{N_1} \) empirically. Diving into the gradient descent process, we discovered the cause of EA. Given the binary classifier with cross-entropy loss \( \mathcal{L}_{\text{BCE}}(x, y) = -y \log p - (1-y) \log (1-p) \) and \texttt{sigmoid} activation function \( \sigma(z) = \frac{1}{1 + e^{-z}} \) at the output layer, the gradient \( \delta \) of loss $\mathcal{L}$ to raw logits \( z \) is \( \delta = \frac{\partial L}{\partial z} = p - y \) according to the chain rule. With a gradient descent step over the imbalanced dataset, the cumulative gradient is computed as:
\begin{equation}
    \label{eq:bin_grad}
    \sum \delta = \underbrace{N_1 (\overline{p^{(1)}} - 1)}_{\text{active gradient}} + \underbrace{N_0 \overline{p^{(0)}}}_{\text{inactive gradient}},
\end{equation}
where \( \overline{p^{(1)}} = \mathbb{E}(p|y=1) \) and \( \overline{p^{(0)}} = \mathbb{E}(p|y=0) \) are the average predictions for positive and negative samples, respectively. In line with Eq. (\ref{eq:type_error_1_n_2}), it is clear the active gradient is \( N_i \) times the Type I error and the inactive gradient is \( N_j \) times the Type II error. As the optimization process aims to locate the global optimum and find the weights with zero gradient, the cumulative gradient in Eq.~(\ref{eq:bin_grad}) always converges to zero. Thus, let Eq.~(\ref{eq:bin_grad}) equals to zero, the EA is calculated as:
\begin{equation}
    \label{eq:bin_ea}
    \mathcal{E} = \frac{1 - \overline{p^{(1)}}}{\overline{p^{(0)}}} \approx \frac{N_0}{N_1}=r.
\end{equation}

\subsection{Bounded Error Asymmetry}

Similar to binary classification with \texttt{sigmoid} activation function and BCE loss, the derivative of loss regarding raw logits in multi-class classification with \texttt{softmax} activation function and cross-entropy is \( \bm{\delta} = \bm{p} - \bm{y} \). This is the same as the binary case, except that \( \bm{\delta}, \bm{p}, \) and \( \bm{y} \) are vectors. Therefore, the above analysis on the gradient and EA can be generalized to multi-class problems. Concretely, given a gradient descent step in multi-class classification, the cumulative gradient is defined by:
\begin{equation}
    \label{eq:mc_grad_onehot}
    \sum_{D} \bm{\delta}_i = N_i \underbrace{(\overline{\bm{p}_i^{(i)}} - 1)}_{u_{\text{I}}} + \underbrace{\sum_{j \neq i} (\overline{\bm{p}_i^{(j)}})}_{u_{\text{II}}} N_j,
\end{equation}
in which \( \overline{p_i^{(j)}} = \frac{1}{N_j} \sum_{y=j} f(x)_i \) is the average probability of predicting data with label \( j \) as class \( i \).

\begin{definition}
    \label{def:ea}
    For a binary classifier trained with sigmoid activation function at the last layer and binary cross-entropy loss, or a multi-class classifier with softmax activation function at the last layer and cross-entropy loss, the EA is the ratio of Type I error to Type II error for the target class $i$:
    \begin{equation}
        \label{eq:ea_def}
        \mathcal{E}(i) = \frac{1 - \overline{p_i^{(i)}}}{\sum_{j \neq i} \overline{p_i^{(j)}}}.
    \end{equation}
\end{definition}

By setting Eq.~(\ref{eq:mc_grad_onehot}) to zero, and simply scaling and rearranging the terms, we can solve for EA in different scenarios. When the dataset is balanced (\(N_i = N_j, \forall i, j\)), the EA of all classes is close to 1. When the dataset is imbalanced (\(N_i \ll N_j, \exists i, \forall j \neq i\)), the EA of the minority class \(i\) is larger than 1:
\begin{equation}
    \begin{aligned}
        \mathcal{E}(i) & = \frac{N_i (1 - \overline{p_i^{(i)}})}{\sum_{j \neq i} N_j \overline{p_i^{(j)}}}   \\
                       & \gg \frac{N_i (1 - \overline{p_i^{(i)}})}{\sum_{j \neq i} N_i \overline{p_i^{(j)}}} \\
                       & = 1.                                                                                \\
    \end{aligned}
\end{equation}

\section*{Gradient Alignment via Label Calibration}
\label{sec:label_calibration}


The model's bias toward the minority class is linked to EA and can be mitigated by aligning gradients to force the EA to 1. Over-sampling and under-sampling methods follow this gradient-based approach by scaling class samples. Re-sampling and re-weighting apply a weight $\alpha_i$ to class $i$, ensuring $\alpha_i N_i = \alpha_j N_j, \forall i, j$. This weighting, applied to Eq.~(\ref{eq:mc_grad_onehot}), eliminates the EA, setting it to 1.



Two limitations exist: re-sampling and re-weighting fail for missing classes, and weighting risks overfitting or underfitting. We address this with class-wise gradient alignment (GA), rescaling inactive gradients to match active ones by adjusting $N_j$ to $N_i$ in Eq.~(\ref{eq:mc_grad_onehot}):

\begin{equation}
    \label{eq:ga_res}
    \sum_{D} \delta_i = N_i \overline{u_{\text{I}}} +
    \sum_{j \neq i} N_j \overline{u_{\text{II}}}.
\end{equation}



Aligned gradients equalize Type I and II errors. For missing class $i$, the inactive gradient is rescaled to zero, preventing forgetting and bias. GA uses calibrated label embedding, replacing one-hot labels $\mathds{1}(\cdot)$ with $\mathbb{C}(\cdot)$. With calibrated label $\bm{q}^{(i)}$, Eq.~(\ref{eq:mc_grad_onehot}) becomes:
\begin{equation}
    \label{eq:mc_grad_cal}
    \sum_{D} \delta_i = N_i (\overline{p_i^{(i)}} - \bm{q}_i^{(i)}) +
    \sum_{j \neq i} N_j (\overline{p_i^{(j)}} - \bm{q}_i^{(j)}).
\end{equation}

To achieve the aligned gradient as in Eq.~(\ref{eq:ga_res}), we have $N_j (\overline{p_i^{(j)}} - \bm{q}_i^{(j)}) = N_i \overline{p_i^{(j)}}$. By solving this, the calibrated label for inactive class $k \neq i$ is:
\begin{equation}
    \label{eq:cal_label}
    \bm{q}_i^{(j)} = \frac{N_j - N_i}{N_j} \overline{p_i^{(j)}}.
\end{equation}

As such, the calibrated label for the imbalanced dataset can be summarized as:
\begin{equation}
    \label{eq:cal}
    \mathbb{C}_i(j) = \begin{cases}
        1                                          & \text{if } j = i, \\
        \frac{N_j - N_i}{N_j} \overline{p_i^{(j)}} & \text{otherwise.}
    \end{cases}
\end{equation}


By combining Eqs.~(\ref{eq:ea_def}), (\ref{eq:mc_grad_cal}), and (\ref{eq:cal}) and setting Eq.(\ref{eq:cal_label}) to zero, GA eliminates EA, except for missing classes where the gradient is zero, preventing bias and class forgetting. Unlike under-sampling, GA mutes only the missing class gradient, preserving learning. FedGA is summarized in Algorithm \ref{alg:GA}.


\section*{Experiments}
\label{sec:experiments_and_results}

To evaluate the proposed GA algorithm with label calibration, we conducted four experiments. Along with FedAvg, we included FedProx~\cite{li2020federated} as a baseline. Three variants of FedAvg, CLIMB, FedRL, and FedNTD, were compared on five benchmark datasets: MNIST~\cite{deng2012mnist}, CIFAR-10, CIFAR-100~\cite{Krizhevsky2009LearningML}, SVHN~\cite{netzer2011reading}, and Tiny-ImageNet~\cite{le2015tiny}, chosen for their varying data heterogeneity and classification difficulty. Each dataset was split among 100 clients, with 10 randomly selected each round. Additional results are in the \emph{supplementary materials}.



\begin{algorithm}[!ht]
    \renewcommand\algorithmicrequire{\textbf{ServerExecute:}}
    \renewcommand\algorithmicensure{\textbf{ClientUpdate:}}
    \renewcommand{\algorithmiccomment}[1]{\hspace{0.1in} $\triangleright$ #1}
    \caption{Federated Learning with Gradient Alignment (FedGA)}
    \label{alg:GA}
    \begin{algorithmic}[1]
        \REQUIRE
        \FOR{round $t \gets$ 0 to $T-1$}
        \STATE select $K$ out of $P$ clients randomly
        \FOR{$k \gets$ 0 to $K-1$ \textbf{synchronously}}
        \STATE $w_{t+1}^k \gets$ ClientUpdate($w_t$)
        \ENDFOR
        \STATE aggregate to $w_{t+1}$ \COMMENT{debiased model aggregation}
        \ENDFOR
    \end{algorithmic}
    \begin{algorithmic}[1]
        \ENSURE
        \STATE load global model $w$
        \FOR{$e \gets 1, E$}
        \FOR{$\bm{x}, \bm{y} \in S^{(k)}$}
        \STATE $\bm{p}=f(\bm{x})$ \COMMENT{batch prediction}
        \STATE $\bm{q} = \mathbb{C}(\bm{y}, \bm{p})$ \COMMENT{batch calibrated label, Eq. (\ref{eq:cal})}
        \STATE $l = \mathcal{L}(\bm{q}, \bm{p})$
        \STATE update $w$ \COMMENT{debiased model update}
        \ENDFOR
        \ENDFOR
        \STATE feedback updated $w$ to server
    \end{algorithmic}
\end{algorithm}

\subsection{Inter-client Imbalance Impact}

\begin{figure}[!ht]
    \centering
    \includegraphics[width=.9\linewidth]{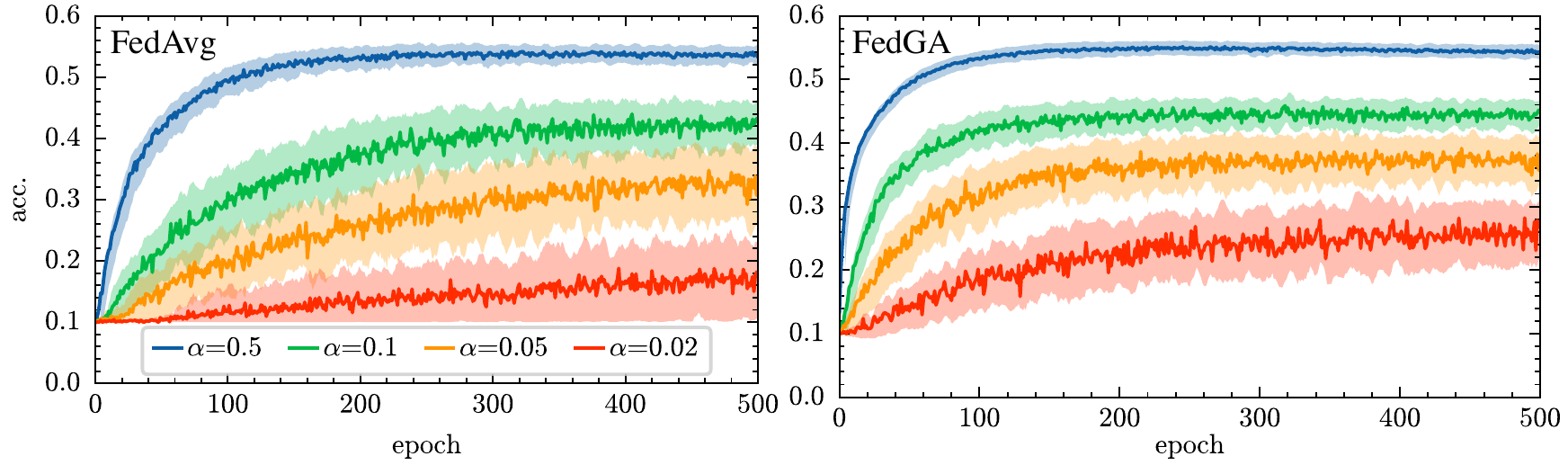}
    \caption{
        Comparative accuracy of FedAvg and FedGA on CIFAR-10 with four inter-client imbalance levels. FedGA converges much faster than FedAvg at each $\alpha$.
    }
    \label{fig:alpha}
\end{figure}

To assess inter-client imbalance in FL, we conducted an ablation study comparing FedGA and FedAvg on client datasets sampled by Dirichlet($\alpha$). Fig. \ref{fig:alpha} reveals that FedGA's accuracy is less impacted by $\alpha$ than FedAvg, where smaller $\alpha$ increases imbalance and reduces performance.

\subsection{Measurable EA Reduction}

\begin{figure}[!ht]
    \centering
    \includegraphics[width=0.5\linewidth]{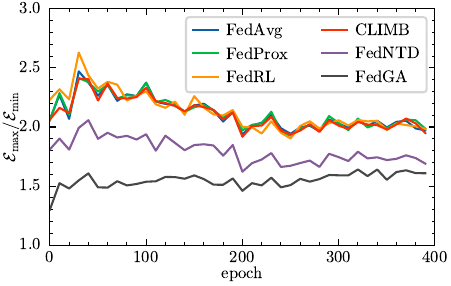}
    \caption{
        The average EA ratio of ten active clients on inter-client imbalance CIFAR-10 during the training phase.
    }
    \label{fig:ea}
\end{figure}

Given Definition~\ref{def:ea}, client EA is measured for all algorithms during training. The EA ratio $\mathcal{E}{\text{max}} / \mathcal{E}{\text{min}}$ tracks class imbalance every ten iterations. Fig. \ref{fig:ea} shows FedGA’s average

\begin{figure*}[!t]
    \centering
    \includegraphics[width=\textwidth]{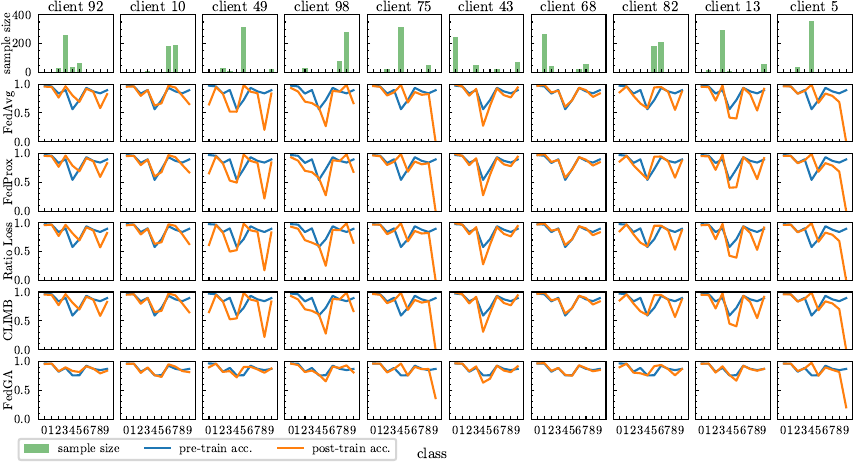}
    \caption{
        Sample distribution as well as pre-train and post-train accuracies of 10 random active clients executing five FL algorithms on inter-client imbalance MNIST dataset. Best viewed in color and zoomed mode.
    }
    \label{fig:pre_post_acc}
\end{figure*}




\noindent EA ratio is consistently lower than others, indicating GA reduces EA, as a ratio near 1 implies balanced EA across classes.

\subsection{Quantifiable Model Bias Reduction}
We measure class-wise accuracy before and after local training to evaluate the bias reduction of the approaches selected for systematic comparisons. The test dataset employed is a standalone balanced dataset. Fig.~\ref{fig:pre_post_acc} illustrates the average accuracy of 10 active clients before and after local training, along with the client sample distribution. Notably, GA exhibits significantly less catastrophic forgetting on the rare and missing classes compared to the other four algorithms. Additionally, the accuracy of GA before local training is less biased than the other algorithms, suggesting that GA aggregates into a more balanced global model in the last iteration.

\subsection{Overall Classification Comparisons}

\begin{table*}[!ht]
    \centering
    \setlength{\tabcolsep}{4pt}
    \begin{tabular}{lllllllll}
        \toprule
        Dataset                            & $\alpha$                 & Met. & FedAvg                  & FedProx       & FedRL                   & CLIMB                   & FedNTD                  & FedGA                   \\
        \midrule
        \multirow[c]{10}{*}{MNIST}         & \multirow[c]{2}{*}{0.05} & F1   & 0.799 (0.020)           & 0.799 (0.020) & 0.837 (0.013)           & 0.803 (0.020)           & 0.810 (0.020)           & \bfseries 0.852 (0.013) \\
                                           &                          & acc. & 0.812 (0.016)           & 0.812 (0.016) & 0.845 (0.011)           & 0.815 (0.016)           & 0.821 (0.018)           & \bfseries 0.859 (0.011) \\
                                           & \multirow[c]{2}{*}{0.1}  & F1   & 0.841 (0.006)           & 0.841 (0.006) & 0.873 (0.008)           & 0.843 (0.006)           & 0.856 (0.004)           & \bfseries 0.880 (0.001) \\
                                           &                          & acc. & 0.849 (0.005)           & 0.849 (0.005) & 0.877 (0.007)           & 0.852 (0.004)           & 0.863 (0.004)           & \bfseries 0.885 (0.001) \\
                                           & \multirow[c]{2}{*}{0.5}  & F1   & 0.904 (0.001)           & 0.903 (0.001) & 0.916 (0.003)           & 0.904 (0.002)           & 0.912 (0.002)           & \bfseries 0.917 (0.001) \\
                                           &                          & acc. & 0.905 (0.002)           & 0.905 (0.002) & 0.917 (0.003)           & 0.906 (0.002)           & 0.914 (0.003)           & \bfseries 0.918 (0.001) \\
                                           & \multirow[c]{2}{*}{1}    & F1   & 0.913 (0.001)           & 0.913 (0.001) & \bfseries 0.924 (0.006) & 0.913 (0.001)           & 0.917 (0.001)           & 0.921 (0.001)           \\
                                           &                          & acc. & 0.914 (0.001)           & 0.914 (0.001) & \bfseries 0.925 (0.006) & 0.914 (0.001)           & 0.919 (0.001)           & 0.922 (0.001)           \\
                                           & \multirow[c]{2}{*}{10}   & F1   & 0.923 (0.001)           & 0.923 (0.001) & \bfseries 0.934 (0.004) & 0.923 (0.001)           & 0.924 (0.001)           & 0.923 (0.001)           \\
                                           &                          & acc. & 0.924 (0.001)           & 0.924 (0.001) & \bfseries 0.934 (0.004) & 0.924 (0.001)           & 0.925 (0.001)           & 0.925 (0.001)           \\
        \hline
        \multirow[c]{10}{*}{CIFAR-10}      & \multirow[c]{2}{*}{0.05} & F1   & 0.428 (0.017)           & 0.424 (0.014) & 0.413 (0.013)           & 0.428 (0.015)           & 0.413 (0.011)           & \bfseries 0.443 (0.006) \\
                                           &                          & acc. & 0.446 (0.016)           & 0.443 (0.011) & 0.427 (0.012)           & 0.446 (0.014)           & 0.430 (0.010)           & \bfseries 0.451 (0.006) \\
                                           & \multirow[c]{2}{*}{0.1}  & F1   & 0.440 (0.022)           & 0.441 (0.021) & 0.426 (0.017)           & 0.445 (0.021)           & 0.448 (0.019)           & \bfseries 0.469 (0.011) \\
                                           &                          & acc. & 0.459 (0.019)           & 0.460 (0.017) & 0.440 (0.015)           & 0.463 (0.017)           & 0.464 (0.015)           & \bfseries 0.477 (0.010) \\
                                           & \multirow[c]{2}{*}{0.5}  & F1   & 0.520 (0.005)           & 0.518 (0.007) & 0.497 (0.017)           & 0.518 (0.004)           & 0.525 (0.003)           & \bfseries 0.529 (0.007) \\
                                           &                          & acc. & 0.525 (0.004)           & 0.523 (0.006) & 0.500 (0.017)           & 0.523 (0.003)           & 0.529 (0.003)           & \bfseries 0.532 (0.006) \\
                                           & \multirow[c]{2}{*}{1}    & F1   & 0.537 (0.017)           & 0.535 (0.016) & 0.479 (0.017)           & 0.536 (0.015)           & 0.534 (0.009)           & \bfseries 0.544 (0.006) \\
                                           &                          & acc. & 0.541 (0.017)           & 0.539 (0.016) & 0.485 (0.016)           & 0.540 (0.014)           & 0.538 (0.010)           & \bfseries 0.547 (0.005) \\
                                           & \multirow[c]{2}{*}{10}   & F1   & 0.559 (0.005)           & 0.559 (0.006) & 0.373 (0.205)           & 0.559 (0.007)           & 0.549 (0.009)           & \bfseries 0.565 (0.002) \\
                                           &                          & acc. & 0.561 (0.005)           & 0.561 (0.006) & 0.397 (0.172)           & 0.561 (0.006)           & 0.551 (0.009)           & \bfseries 0.567 (0.002) \\
        \hline
        \multirow[c]{10}{*}{CIFAR-100}     & \multirow[c]{2}{*}{0.05} & F1   & 0.135 (0.016)           & 0.157 (0.018) & \bfseries 0.197 (0.021) & 0.117 (0.009)           & 0.162 (0.022)           & 0.154 (0.018)           \\
                                           &                          & acc. & 0.149 (0.017)           & 0.175 (0.021) & \bfseries 0.211 (0.023) & 0.131 (0.010)           & 0.180 (0.025)           & 0.167 (0.018)           \\
                                           & \multirow[c]{2}{*}{0.1}  & F1   & 0.207 (0.002)           & 0.208 (0.004) & 0.203 (0.018)           & 0.197 (0.007)           & 0.224 (0.011)           & \bfseries 0.227 (0.007) \\
                                           &                          & acc. & 0.223 (0.003)           & 0.225 (0.005) & 0.220 (0.016)           & 0.212 (0.009)           & 0.237 (0.013)           & \bfseries 0.241 (0.007) \\
                                           & \multirow[c]{2}{*}{0.5}  & F1   & 0.294 (0.003)           & 0.297 (0.002) & 0.197 (0.006)           & 0.294 (0.003)           & \bfseries 0.299 (0.006) & 0.299 (0.003)           \\
                                           &                          & acc. & 0.304 (0.003)           & 0.306 (0.001) & 0.222 (0.006)           & 0.304 (0.003)           & \bfseries 0.309 (0.004) & 0.306 (0.003)           \\
                                           & \multirow[c]{2}{*}{1}    & F1   & 0.323 (0.008)           & 0.321 (0.008) & 0.205 (0.002)           & 0.322 (0.006)           & 0.318 (0.008)           & \bfseries 0.327 (0.008) \\
                                           &                          & acc. & 0.329 (0.008)           & 0.327 (0.010) & 0.225 (0.004)           & 0.328 (0.008)           & 0.323 (0.008)           & \bfseries 0.330 (0.009) \\
                                           & \multirow[c]{2}{*}{10}   & F1   & 0.339 (0.013)           & 0.336 (0.008) & 0.177 (0.007)           & 0.334 (0.013)           & 0.331 (0.003)           & \bfseries 0.342 (0.005) \\
                                           &                          & acc. & 0.346 (0.012)           & 0.343 (0.007) & 0.198 (0.009)           & 0.342 (0.011)           & 0.340 (0.004)           & \bfseries 0.347 (0.006) \\
        \hline
        \multirow[c]{10}{*}{SVHN}          & \multirow[c]{2}{*}{0.05} & F1   & 0.760 (0.076)           & 0.772 (0.057) & 0.794 (0.023)           & 0.770 (0.050)           & 0.766 (0.053)           & \bfseries 0.801 (0.007) \\
                                           &                          & acc. & 0.776 (0.071)           & 0.787 (0.054) & 0.808 (0.023)           & 0.785 (0.048)           & 0.782 (0.052)           & \bfseries 0.815 (0.008) \\
                                           & \multirow[c]{2}{*}{0.1}  & F1   & 0.817 (0.015)           & 0.817 (0.014) & 0.738 (0.252)           & 0.818 (0.014)           & 0.814 (0.010)           & \bfseries 0.820 (0.008) \\
                                           &                          & acc. & 0.831 (0.013)           & 0.831 (0.013) & 0.758 (0.228)           & 0.831 (0.012)           & 0.828 (0.009)           & \bfseries 0.834 (0.007) \\
                                           & \multirow[c]{2}{*}{0.5}  & F1   & 0.832 (0.009)           & 0.667 (0.365) & 0.494 (0.371)           & 0.673 (0.365)           & 0.829 (0.012)           & \bfseries 0.838 (0.005) \\
                                           &                          & acc. & 0.844 (0.008)           & 0.690 (0.341) & 0.531 (0.342)           & 0.696 (0.341)           & 0.842 (0.010)           & \bfseries 0.851 (0.005) \\
                                           & \multirow[c]{2}{*}{1}    & F1   & \bfseries 0.843 (0.011) & 0.842 (0.007) & 0.424 (0.362)           & 0.840 (0.010)           & 0.839 (0.005)           & 0.840 (0.009)           \\
                                           &                          & acc. & \bfseries 0.856 (0.010) & 0.854 (0.007) & 0.469 (0.329)           & 0.852 (0.009)           & 0.853 (0.005)           & 0.853 (0.009)           \\
                                           & \multirow[c]{2}{*}{10}   & F1   & 0.848 (0.006)           & 0.848 (0.006) & 0.018 (0.002)           & \bfseries 0.850 (0.006) & 0.844 (0.009)           & 0.848 (0.007)           \\
                                           &                          & acc. & 0.860 (0.006)           & 0.861 (0.006) & 0.102 (0.011)           & \bfseries 0.862 (0.005) & 0.857 (0.008)           & 0.861 (0.007)           \\
        \hline
        \multirow[c]{10}{*}{Tiny-ImageNet} & \multirow[c]{2}{*}{0.05} & F1   & 0.055 (0.015)           & 0.085 (0.010) & \bfseries 0.091 (0.011) & 0.043 (0.013)           & 0.084 (0.008)           & 0.065 (0.015)           \\
                                           &                          & acc. & 0.075 (0.015)           & 0.106 (0.010) & \bfseries 0.110 (0.007) & 0.057 (0.013)           & 0.103 (0.008)           & 0.086 (0.015)           \\
                                           & \multirow[c]{2}{*}{0.1}  & F1   & 0.114 (0.011)           & 0.119 (0.003) & 0.102 (0.024)           & 0.103 (0.023)           & 0.123 (0.017)           & \bfseries 0.124 (0.008) \\
                                           &                          & acc. & 0.134 (0.011)           & 0.142 (0.002) & 0.120 (0.024)           & 0.120 (0.024)           & 0.139 (0.016)           & \bfseries 0.144 (0.007) \\
                                           & \multirow[c]{2}{*}{0.5}  & F1   & 0.174 (0.016)           & 0.179 (0.014) & 0.052 (0.020)           & 0.177 (0.010)           & \bfseries 0.195 (0.006) & 0.193 (0.012)           \\
                                           &                          & acc. & 0.195 (0.015)           & 0.199 (0.012) & 0.077 (0.020)           & 0.196 (0.008)           & \bfseries 0.212 (0.004) & 0.209 (0.012)           \\
                                           & \multirow[c]{2}{*}{1}    & F1   & 0.206 (0.013)           & 0.205 (0.009) & 0.039 (0.016)           & 0.201 (0.009)           & 0.210 (0.010)           & \bfseries 0.213 (0.009) \\
                                           &                          & acc. & 0.222 (0.014)           & 0.221 (0.010) & 0.061 (0.017)           & 0.216 (0.009)           & 0.223 (0.011)           & \bfseries 0.225 (0.009) \\
                                           & \multirow[c]{2}{*}{10}   & F1   & 0.236 (0.007)           & 0.235 (0.005) & 0.053 (0.015)           & 0.224 (0.007)           & 0.239 (0.004)           & \bfseries 0.243 (0.004) \\
                                           &                          & acc. & 0.247 (0.008)           & 0.246 (0.004) & 0.079 (0.016)           & 0.235 (0.007)           & 0.249 (0.005)           & \bfseries 0.251 (0.005) \\
        \bottomrule
    \end{tabular}
    \caption{Validation accuracy and F1 score of different FL algorithms on MNIST, CIFAR-10, CIFAR-100, SVHN and Tiny-ImageNet with different levels of heterogeneity. The results are averaged over five independent runs with standard deviation reported in parentheses. The best results for each setting $\alpha$ are marked in bold. }
    \label{tab:nes_results}
\end{table*}




A balanced test dataset is used to evaluate the performance of the selected FL re-balancing algorithms, ensuring fair assessment across all classes. Overall accuracy remains a key criterion for model selection, while the multi-class F1 score is calculated to provide a more comprehensive evaluation of performance across imbalanced classes. To ensure statistical reliability, each experiment is repeated 20 times with random seeds (from 0 to 19), reducing the influence of variability and ensuring consistent results \cite{wang2024unified}.


The average accuracy and F1 score, along with the corresponding standard deviation, are detailed in Tab.\ref{tab:nes_results}. Across experiments conducted on various datasets and under different levels of class imbalance, GA consistently demonstrates superior performance in both accuracy and F1 score compared to the other four algorithms. This highlights the effectiveness of GA in addressing data heterogeneity and maintaining robust performance across diverse scenarios. Additionally, the smaller standard deviation observed in GA’s results indicates greater stability and reliability, underscoring its ability to produce consistent outcomes across multiple runs. Notably, for the SVHN dataset\cite{netzer2011reading} with $\alpha = 0.1$ and $0.5$, GA exhibits a slightly larger standard deviation. This deviation, however, reflects the instability of the baseline algorithms, which fail to converge within the allocated number of global iterations, rather than a limitation of GA. This reinforces GA’s resilience in challenging conditions where other algorithms struggle to achieve convergence.

\begin{figure}[htbp]
    \centering
    \includegraphics[width=\linewidth]{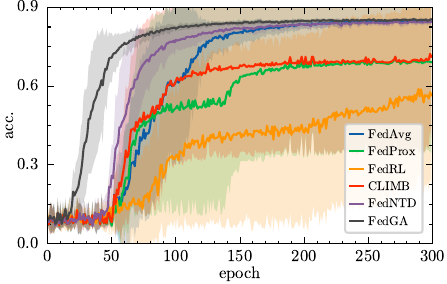}
    \caption{
        The accuracy curve of GA and other algorithms on an inter-client imbalanced SVHN dataset sampled with $\alpha = 0.5$.
        The shaded area denotes the standard deviation over five independent runs.
    }
    \label{fig:cifar10}
\end{figure}

On the simple MNIST~\cite{deng2012mnist} dataset, GA's performance is slightly better than that of other algorithms. As $\alpha$ decreases from 0.5 to 0.05, the inter-client imbalance level increases. The performance advantage of GA over other algorithms expands with the increased inter-client imbalance. For the challenging tasks of SVHN~\cite{netzer2011reading} and CIFAR-10~\cite{Krizhevsky2009LearningML}, GA significantly outperforms the other algorithms. For instance, GA achieves 72\% accuracy on SVHN with $\alpha=0.1$, while the others remain below 20\%. Even when $\alpha=0.05$, GA still achieves 28.9\% accuracy on SVHN, outperforming the others, which hover around 10\%, akin to random guessing.
In addition to the average accuracy and F1 score, the convergence speed stands out as a strength of GA. As illustrated in Fig.~\ref{fig:cifar10}, GA converges significantly faster than the other algorithms. When considering the time it takes to reach 90\% of the final accuracy, GA is three times faster than the other algorithms in the selected experiment.

\section*{Conclusion}
\label{sec:conclusion}

This paper investigates the model bias caused by FL inter-
client imbalance from a novel perspective of the gradient. It
identifies the asymmetric Type I and Type II errors in class
imbalance and establishes a connection between the gradient of loss and the raw logits. The proposed GA approach, \ie FedGA, involves label calibration to align active and inactive gradients, ultimately eliminating the EA. Extensive experimental results confirm the efficacy and efficiency of the FedGA in eliminating EA and reducing client bias. Experimental results also inspire further investigations on the optimal sample selection \cite{cui2019class} for more robust label calibration, as well as GA extension for a wider spectrum of downstream tasks.


\section*{Acknowledgements}
This work was supported by the EPSRC Early Career Researchers International Collaboration Grants (EP/Y002539/1) and the National Natural Science Foundation for Young Scientists of China [62206239].
The computations described in this research were performed using the Baskerville Tier 2 HPC service (https://www.baskerville.ac.uk/).
Baskerville was funded by the EPSRC and UKRI through the World Class Labs scheme (EP/T022221/1) and the Digital Research Infrastructure programme (EP/W032244/1) and is operated by Advanced Research Computing at the University of Birmingham.
Chenguang Xiao is partially supported by the Chinese Scholarship Council.
\appendix

\bibliography{aaai25}

\end{document}


\maketitle

%
\section{Algorithm Details}

\subsection{Calibrated Label Calculation}

The calibrated label, calculated by Eq.~(10) in the main body of our paper, is used to measure the loss during client training. To improve computational efficiency, we calculate the calibrated label directly from each sample's output $\bm{p^{(i)}}$, rather than using the average prediction $\overline{\bm{p^{(i)}}}$ of class $i$ samples for all data. This adjustment does not impact the analysis of EA or gradient alignment, ensuring the validity of our approach while reducing computational overhead.

Here is an example illustrating the calculation of the calibrated label $\bm{q}$ during a training step. Consider a classification task with labels in the range $[0, 4]$, where the label distribution is represented by the vector $\bm{y} = [100, 0, 20, 5, 20]$. During a mini-batch gradient descent step with data $\xi$, the model predicts $\bm{p} = [0.8, 0.1, 0.05, 0.01, 0.04]$ for a sample from class 0. The calibrated label $\bm{q}$ is then derived based on this prediction and the label distribution as follows:

\begin{equation}
    \begin{aligned}
        \bm{q}  = & \mathbb{C}(\bm{y}, \bm{p})                                           \\
        =         & \Big[1, \frac{100-0}{100}\times 0.1, \frac{100-20}{100} \times 0.05, \\
                  & \frac{100-5}{100} \times 0.01, \frac{100-20}{100} \times 0.04\Big]   \\
        =         & [1, 0.1, 0.04, 0.095, 0.08].
    \end{aligned}
\end{equation}


The calibrated label $\bm{q}$ adjusts the standard one-hot label $[1, 0, 0, 0, 0]$ by penalizing classes with fewer data samples. This reweighting helps address class imbalance, ensuring that underrepresented classes receive greater influence during training, ultimately improving model performance and reducing bias.

\section{Experimental Details}
\label{app:experimental_details}

Details of the experimental setup are provided in this section, including the FL system information, simulation of inter-client imbalance, model architecture, hyperparameters fine-tuning, computing resources and learning platforms used in the experiments.

All the experiments are conducted on a FL system with 100 clients, and 10 active clients are selected randomly in each round.
The global epoch is set according to the difficulty of the task, specifically minimal iterations for FedAvg to converge on the task.
The local epoch is set to 2 for the experiments mentioned in the main paper.
Experiments with more local epochs are included in section~\ref{sec:sensitive_to_local_epoch} to show the sensitivity of GA to the local epoch.


\subsection{Datasets}
Three real-world datasets with increasing classification challenges are used in the experiments: MNIST, SVHN, and CIFAR-10. MNIST~\cite{deng2012mnist} is a handwritten digit dataset with 10 classes, 60,000 training samples, and 10,000 test samples of size (1, 28, 28). SVHN~\cite{netzer2011reading} is a dataset of house numbers obtained from Google Street View images, containing 73,257 training images and 26,032 test images of size (3, 32, 32) with 10 classes. CIFAR-10~\cite{krizhevsky2009learningML} contains 50,000 training images and 10,000 test images of size (3, 32, 32), and consists of natural images categorized into 10 classes.

Datasets mentioned above contains only 10 classes, which are relatively easy for the FL system to learn.
To simulate more complex scenarios, CIFAR-100 and Tiny-ImageNet are also included in the experiments.
CIFAR-100~\cite{krizhevsky2009learningML} is a dataset with 100 classes, 50,000 training images, and 10,000 test images of size (3, 32, 32).
Tiny-ImageNet~\cite{le2015tiny} is a subset of the ImageNet dataset with 200 classes, 100,000 training images, and 10,000 test images of size (3, 64, 64).
There are a few images in Tiny-ImageNet that are greyscale, which are converted to RGB by duplicating the greyscale channel to three channels throughout the experiments.
With more classes, CIFAR-100 and Tiny-ImageNet are more challenges in terms of heterogeneity and imbalance.

\subsection{Imbalance Simulation}
To simulate the inter-client imbalance, the label distribution of each client is sampled from a Dirichlet distribution $\text{Dir}(\alpha)$ with $\alpha$ values in \{10, 1, 0.5, 0.1, 0.05\}. The smaller the $\alpha$, the more severe the inter-client imbalance. Class missing is prevalent in the Dirichlet distribution with the selected $\alpha$ values.



\begin{figure}[htbp]
    \centering
    \includegraphics[width=0.9\linewidth]{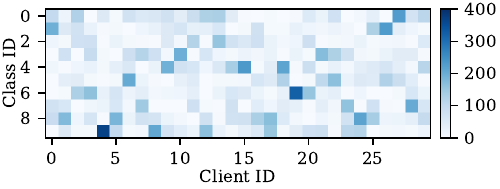}
    \caption{Local label distribution for first 30 out of 100 clients with $\alpha=0.5$ in a 10 class classification task.}
    \label{fig:hetero-0.5}
\end{figure}

\begin{figure}[htbp]
    \centering
    \includegraphics[width=.9\linewidth]{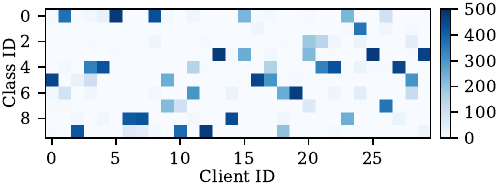}
    \caption{Local label distribution for first 30 out of 100 clients with $\alpha=0.05$ in a 10 class classification task.}
    \label{fig:hetero-0.05}
\end{figure}

Figure~\ref{fig:hetero-0.5} and Figure~\ref{fig:hetero-0.05} show the local label distribution for the first 30 out of 100 clients with $\alpha=0.5$ and $\alpha=0.05$ in a 10 class classification task, respectively.
Each column represents the label distribution of one client, and the colour of the cell represents the number of samples in the corresponding class.
Darker colours indicate more samples in the class.

It can be observed that the label distribution of clients becomes more imbalanced as $\alpha$ decreases.
When $\alpha=0.05$, most clients have their data concentrated in a 1 or 2 classes, which is more challenging for the FL system to learn.
For example, most data of client 0 comes from class 3 and 6 while most data of client 1 comes from class 1, resulting in different objectives for the 2 clients.
The mini-batch size is set to 64 for all the experiments during the local training process.

\subsection{Model Architecture}
Three different type of neural net works are adopted in the experiments to evaluate the robustness of GA in FL.
From simple MLP, to classic CNN and advanced ResNet, models used in the experiments cover a wide range of neural network architectures.
This also fits to different tasks with varying difficulty levels in the experiments.

For the easy task MNIST, a simple 3-layer MLP with a hidden layer of size 128 is used.
The activation function is ReLU for the first two layers.

For the medium task SVHN and CIFAR-10, a LeNet with 3 input channels, max pooling layers, and ReLU activations is used.

For the hard task CIFAR-100 and Tiny-ImageNet, a 20 layer ResNet is used.
This ResNet implementation for CIFAR-10 and CIFAR-100 is based on https://github.com/akamaster/pytorch\_resnet\_cifar10.
As other popular ResNet implementation for CIFAR-10 failed to follow the original paper, we choose this implementation for fair comparison.

\subsection{Hyperparameters Fine-Tuning}
For the MNIST, SVHN, and CIFAR-10 datasets, the SGD optimizer with a momentum of 0.9 is used.
For each heterogeneity levels, the learning rate is grid searched in range of $\{1, 0.1, 0.01, 0.001, 0.0001\}$ based on FedAvg.
Then same learning rate is used for all other FL algorithms in same task.

Apart from SGD, Adam is another popular optimizer wildly used in FL.
Although most of the existing FL algorithms are analysed with SGD, we also include Adam in the experiments to show the robustness of GA in FL.
Adam optimizer includes there hyperparameters: learning rate, $\beta_1$, and $\beta_2$.
We keep $\beta_1=0.9$ and $\beta_2=0.999$ fixed and grid search the learning rate in range of $\{1, 0.1, 0.01, 0.001, 0.0001\}$ for each task using FedAvg.
Then same learning rate is used for all other FL algorithms in same task.
Table~\ref{tab:hyperparameters} shows the best hyperparameters used in the experiments after above grid search.

\begin{table}[htbp]
    \centering
    \begin{tabular}{lrlrlrr}
        \toprule
        Dataset                        & Model                   & $\alpha$ & Epoch & LR  \\
        \midrule
        \multirow{5}{*}{MNIST}         & \multirow{5}{*}{MLP}    & 10       & 100   & 0.1 \\
                                       &                         & 1        & 100   & 0.1 \\
                                       &                         & 0.5      & 100   & 0.1 \\
                                       &                         & 0.1      & 100   & 0.1 \\
                                       &                         & 0.05     & 100   & 0.1 \\
        \hline
        \multirow{5}{*}{CIFAR-10}      & \multirow{5}{*}{LeNet}  & 10       & 300   & 0.1 \\
                                       &                         & 1        & 300   & 0.1 \\
                                       &                         & 0.5      & 400   & 0.1 \\
                                       &                         & 0.1      & 400   & 0.1 \\
                                       &                         & 0.05     & 600   & 0.1 \\
        \hline
        \multirow{5}{*}{CIFAR-100}     & \multirow{5}{*}{ResNet} & 10       & 500   & 0.1 \\
                                       &                         & 1        & 500   & 0.1 \\
                                       &                         & 0.5      & 600   & 0.1 \\
                                       &                         & 0.1      & 700   & 0.1 \\
                                       &                         & 0.05     & 900   & 0.1 \\
        \hline
        \multirow{5}{*}{SVHN}          & \multirow{5}{*}{LeNet}  & 10       & 200   & 0.3 \\
                                       &                         & 1        & 200   & 0.3 \\
                                       &                         & 0.5      & 300   & 0.3 \\
                                       &                         & 0.1      & 600   & 0.1 \\
                                       &                         & 0.05     & 600   & 0.1 \\
        \hline
        \multirow{5}{*}{Tiny-ImageNet} & \multirow{5}{*}{ResNet} & 10       & 500   & 0.3 \\
                                       &                         & 1        & 500   & 0.3 \\
                                       &                         & 0.5      & 600   & 0.3 \\
                                       &                         & 0.1      & 900   & 0.3 \\
                                       &                         & 0.05     & 1100  & 0.3 \\
        \bottomrule
    \end{tabular}
    \caption{Hyperparameters used in the experiments based on grid search with FedAvg algorithm given the dataset and Dirichlet distribution parameter $\alpha$.}
    \label{tab:hyperparameters}
\end{table}

Some comparison algorithms require additional hyperparameters to be set.
For FedProx, following the original papers to grid search in range of $\{1, 0.1, 0.01, 0.001\}$ for the proximal weight.
The best proximal term is all 0.001 for all the tasks.

For CLIMB, the tolerance parameter $\epsilon$ is grid searched in range of $\{1, 0.1, 0.01, 0.001\}$ and dual learning rate $\eta_d$ is grid searched in range of $\{4, 2, 1, 0.5, 0.1, 0.05\}$.
The best $\epsilon$ and $\eta$ are shown in Table~\ref{tab:hyperparameters_climb} for each dataset.
\begin{table}[htbp]
    \centering
    \setlength{\tabcolsep}{16pt}
    \begin{tabular}{lrr}
        \toprule
        Dataset       & $\epsilon$ & $\eta_d$ \\
        \midrule
        MNIST         & 2          & 0.01     \\
        CIFAR-10      & 0.1        & 0.1      \\
        CIFAR-100     & 0.05       & 1        \\
        SVHN          & 0.5        & 0.1      \\
        Tiny-ImageNet & 0.1        & 0.1      \\
        \bottomrule
    \end{tabular}
    \caption{CLIMB hyperparameters used in the experiments based on grid search with FedAvg algorithm for different datasets.}
    \label{tab:hyperparameters_climb}
\end{table}

FedNTD~\cite{leePreservationGlobalKnowledge2022} preserves the global perspective on locally available data only for the not-true classes.
We use the recommended hyperparameters same as the original paper, which is $\tau=1$ and $\beta=1$ for all the tasks.

\subsection{Computing Resources and Learning Platforms}
The experiments presented in this work were conducted on an Intel Xeon Platinum 8360Y CPU with 64 GB memory and an NVIDIA A100 GPU with 40 GB memory. The Operating System is Red Hat Enterprise 8.9. The average running time of 1000 iterations for MNIST, SVHN, and CIFAR-10 is 53 minutes, 1 hour, and 1 hour 13 minutes, respectively.

We implement all the FL algorithms based on PyTorch version 2.0.1 and Python version 3.11.9.
Wandb is used to log the training process and gather the results.
For detailed package usage, please refer to the requirement in our code repository.

\section{Additional Experiment Results}

The results are presented by mean and standard deviation of accuracy and F1 score over 5 runs with random seed from 0 to 4.
The scores are rounded to 3 decimal places for better readability.
Table~\ref{tab:nes_results_80} shows the prediction scores of different FL algorithms on CIFAR-10, MNIST, and SVHN at 80\% of the original global iteration.
E.g., the scores are measured at 100 epochs for MNIST dataset with $\alpha=10$ in the main paper.
Then the scores measured at $100 \times 0.8 = 80$ epochs are shown in Table~\ref{tab:nes_results_80}.
Scores in the training stage better reflect speed and stability of the FL algorithms, which are important for practical applications.

\begin{table*}[htbp]
    \centering
    \renewcommand{\arraystretch}{1.4}
    \begin{tabular}{llllllll}
        \toprule
        Dataset                       & $\alpha$                 & Met. & FedAvg                  & FedProx       & CLIMB                   & FedNTD                  & FedGA                   \\
        \midrule
        \multirow[c]{10}{*}{MNIST}    & \multirow[c]{2}{*}{0.05} & F1  & 0.873 (0.012)           & 0.873 (0.012) & 0.876 (0.012)           & 0.880 (0.017)           & \bfseries 0.898 (0.007) \\
                                      &                          & Acc. & 0.876 (0.011)           & 0.876 (0.011) & 0.879 (0.012)           & 0.882 (0.016)           & \bfseries 0.900 (0.007) \\
                                      & \multirow[c]{2}{*}{0.1}  & F1  & 0.891 (0.005)           & 0.891 (0.005) & 0.895 (0.004)           & 0.905 (0.003)           & \bfseries 0.913 (0.002) \\
                                      &                          & Acc. & 0.893 (0.004)           & 0.893 (0.004) & 0.896 (0.004)           & 0.907 (0.003)           & \bfseries 0.915 (0.002) \\
                                      & \multirow[c]{2}{*}{0.5}  & F1  & 0.926 (0.001)           & 0.926 (0.001) & 0.927 (0.002)           & \bfseries 0.936 (0.002) & \bfseries 0.936 (0.002) \\
                                      &                          & Acc. & 0.927 (0.001)           & 0.927 (0.001) & 0.928 (0.002)           & 0.936 (0.002)           & \bfseries 0.937 (0.002) \\
                                      & \multirow[c]{2}{*}{1}    & F1  & 0.934 (0.002)           & 0.934 (0.002) & 0.934 (0.002)           & 0.939 (0.002)           & \bfseries 0.940 (0.001) \\
                                      &                          & Acc. & 0.935 (0.002)           & 0.935 (0.002) & 0.935 (0.002)           & 0.940 (0.002)           & \bfseries 0.940 (0.001) \\
                                      & \multirow[c]{2}{*}{10}   & F1  & 0.942 (0.001)           & 0.942 (0.001) & 0.942 (0.001)           & \bfseries 0.943 (0.001) & 0.942 (0.001)           \\
                                      &                          & Acc. & 0.942 (0.001)           & 0.942 (0.001) & 0.942 (0.001)           & \bfseries 0.944 (0.001) & 0.942 (0.001)           \\
        \midrule
        \multirow[c]{10}{*}{CIFAR-10} & \multirow[c]{2}{*}{0.05} & F1  & 0.428 (0.017)           & 0.424 (0.014) & 0.428 (0.015)           & 0.413 (0.011)           & \bfseries 0.443 (0.006) \\
                                      &                          & Acc. & 0.446 (0.016)           & 0.443 (0.011) & 0.446 (0.014)           & 0.430 (0.010)           & \bfseries 0.451 (0.006) \\
                                      & \multirow[c]{2}{*}{0.1}  & F1  & 0.440 (0.022)           & 0.441 (0.021) & 0.445 (0.021)           & 0.448 (0.019)           & \bfseries 0.469 (0.011) \\
                                      &                          & Acc. & 0.459 (0.019)           & 0.460 (0.017) & 0.463 (0.017)           & 0.464 (0.015)           & \bfseries 0.477 (0.010) \\
                                      & \multirow[c]{2}{*}{0.5}  & F1  & 0.520 (0.005)           & 0.518 (0.007) & 0.518 (0.004)           & 0.525 (0.003)           & \bfseries 0.529 (0.007) \\
                                      &                          & Acc. & 0.525 (0.004)           & 0.523 (0.006) & 0.523 (0.003)           & 0.529 (0.003)           & \bfseries 0.532 (0.006) \\
                                      & \multirow[c]{2}{*}{1}    & F1  & 0.537 (0.017)           & 0.535 (0.016) & 0.536 (0.015)           & 0.534 (0.009)           & \bfseries 0.544 (0.006) \\
                                      &                          & Acc. & 0.541 (0.017)           & 0.539 (0.016) & 0.540 (0.014)           & 0.538 (0.010)           & \bfseries 0.547 (0.005) \\
                                      & \multirow[c]{2}{*}{10}   & F1  & 0.559 (0.005)           & 0.559 (0.006) & 0.559 (0.007)           & 0.549 (0.009)           & \bfseries 0.565 (0.002) \\
                                      &                          & Acc. & 0.561 (0.005)           & 0.561 (0.006) & 0.561 (0.006)           & 0.551 (0.009)           & \bfseries 0.567 (0.002) \\
        \midrule
        
        \multirow[c]{10}{*}{SVHN}     & \multirow[c]{2}{*}{0.05} & F1  & 0.760 (0.076)           & 0.772 (0.057) & 0.770 (0.050)           & 0.766 (0.053)           & \bfseries 0.801 (0.007) \\
                                      &                          & Acc. & 0.776 (0.071)           & 0.787 (0.054) & 0.785 (0.048)           & 0.782 (0.052)           & \bfseries 0.815 (0.008) \\
                                      & \multirow[c]{2}{*}{0.1}  & F1  & 0.817 (0.015)           & 0.817 (0.014) & 0.818 (0.014)           & 0.814 (0.010)           & \bfseries 0.820 (0.008) \\
                                      &                          & Acc. & 0.831 (0.013)           & 0.831 (0.013) & 0.831 (0.012)           & 0.828 (0.009)           & \bfseries 0.834 (0.007) \\
                                      & \multirow[c]{2}{*}{0.5}  & F1  & 0.832 (0.009)           & 0.667 (0.365) & 0.673 (0.365)           & 0.829 (0.012)           & \bfseries 0.838 (0.005) \\
                                      &                          & Acc. & 0.844 (0.008)           & 0.690 (0.341) & 0.696 (0.341)           & 0.842 (0.010)           & \bfseries 0.851 (0.005) \\
                                      & \multirow[c]{2}{*}{1}    & F1  & \bfseries 0.843 (0.011) & 0.842 (0.007) & 0.840 (0.010)           & 0.839 (0.005)           & 0.840 (0.009)           \\
                                      &                          & Acc. & \bfseries 0.856 (0.010) & 0.854 (0.007) & 0.852 (0.009)           & 0.853 (0.005)           & 0.853 (0.009)           \\
                                      & \multirow[c]{2}{*}{10}   & F1  & 0.848 (0.006)           & 0.848 (0.006) & \bfseries 0.850 (0.006) & 0.844 (0.009)           & 0.848 (0.007)           \\
                                      &                          & Acc. & 0.860 (0.006)           & 0.861 (0.006) & \bfseries 0.862 (0.005) & 0.857 (0.008)           & 0.861 (0.007)           \\
        \bottomrule
    \end{tabular}
    \caption{Performance of FL algorithms on different heterogeneity levels.
        The results are averaged over 5 runs with standard deviation in parentheses.
        The best results for each setting are highlighted in bold.
        Results are based on 80\% of the global iterations.
    }
    \label{tab:nes_results_80}
\end{table*}

As shown in Table~\ref{tab:nes_results_80}, FedGA achieves the best performance in terms of accuracy and F1 score in most cases.
It wins all 5 scenarios in CIFAR-10, 8 out of 10 scenarios in MNIST, and 6 out of 10 scenarios in SVHN.
Notably, MNIST is simple dataset with $\alpha=10$ where all selected algorithms already converge to the best performance in 80\% of the global iterations.
If we zoom in to the first 10 iterations, FedGA is still faster than other algorithms in terms of convergence speed as shown in Figure~\ref{fig:mnist-10}.

\begin{figure}[!ht]
    \centering
    \includegraphics[width=.9\linewidth]{fig/sgd_MNIST_10_V-Acc_xlim_10_ylim_(0.5, 0.9).pdf}
    \caption{First 10 iterations accuracy on MNIST dataset with $\alpha=10$.}
    \label{fig:mnist-10}
\end{figure}

Similar results can be observed in SVHN with $\alpha=10, 1$ as shown in Figure~\ref{fig:svhn-10} and Figure~\ref{fig:svhn-1}.
\begin{figure}[!ht]
    \centering
    \includegraphics[width=.9\linewidth]{fig/sgd_SVHN_10_V-Acc_xlim_100_ylim_(0, 0.9).pdf}
    \caption{First 100 iterations accuracy on SVHN dataset with $\alpha=10$.}
    \label{fig:svhn-10}
\end{figure}
\begin{figure}[!ht]
    \centering
    \includegraphics[width=.9\linewidth]{fig/sgd_SVHN_1_V-Acc_xlim_100_ylim_(0, 1).pdf}
    \caption{First 100 iterations accuracy on SVHN dataset with $\alpha=1$.}
    \label{fig:svhn-1}
\end{figure}

To illustrate the convergence speed of selected algorithms in heterogeneity, we also list the global iterations required to reach 90\% of the best performance of FedAvg in Table~\ref{tab:nes_results_80}.
E.g., the first row indicate FedGA spends 67 global iterations to reach 90\% of the accuracy FedAvg achieves in 100 global iterations on CIFAR-10 with $\alpha=10$.
The fewer global iterations required, the faster the algorithm converges to the best performance.
The speed-up factor is included in the brackets for better comparison, which is calculated as the ratio of the global iterations required by FedAvg and the selected algorithm.

Except for the easy task MNIST with $\alpha=10$, FedGA achieves the fastest convergence speed in all the cases.
As explained previously, we can still observe the advance of FedGA on MNIST with $\alpha=10$ in the first 10 iterations as shown in Figure~\ref{fig:mnist-10}.

\begin{table*}[htbp]
    \centering
    \begin{tabular}{llrrrrr}
        \toprule
        Dataset                      & $\alpha$ & FedAvg                       & FedProx                      & CLIMB                        & FedNTD                       & FedGA                        \\
        \midrule
        \multirow[c]{5}{*}{MNIST}    & 10       & \bfseries   1 (1.00$\times$) & \bfseries   1 (1.00$\times$) & \bfseries   1 (1.00$\times$) & \bfseries   1 (1.00$\times$) & \bfseries   1 (1.00$\times$) \\
                                     & 1        & 2 (1.00$\times$)             & 2 (1.00$\times$)             & 2 (1.00$\times$)             & 2 (1.00$\times$)             & \bfseries   1 (2.00$\times$) \\
                                     & 0.5      & 4 (1.00$\times$)             & 4 (1.00$\times$)             & 4 (1.00$\times$)             & 3 (1.33$\times$)             & \bfseries   2 (2.00$\times$) \\
                                     & 0.1      & 9 (1.00$\times$)             & 9 (1.00$\times$)             & 9 (1.00$\times$)             & 8 (1.12$\times$)             & \bfseries   5 (1.80$\times$) \\
                                     & 0.05     & 14 (1.00$\times$)            & 14 (1.00$\times$)            & 14 (1.00$\times$)            & 13 (1.08$\times$)            & \bfseries   8 (1.75$\times$) \\
        \midrule
        \multirow[c]{5}{*}{CIFAR-10} & 10       & 73 (1.00$\times$)            & 74 (0.99$\times$)            & 74 (0.99$\times$)            & 82 (0.89$\times$)            & \bfseries  67 (1.09$\times$) \\
                                     & 1        & 84 (1.00$\times$)            & 83 (1.01$\times$)            & 83 (1.01$\times$)            & 84 (1.00$\times$)            & \bfseries  72 (1.17$\times$) \\
                                     & 0.5      & 97 (1.00$\times$)            & 98 (0.99$\times$)            & 97 (1.00$\times$)            & 96 (1.01$\times$)            & \bfseries  78 (1.24$\times$) \\
                                     & 0.1      & 156 (1.00$\times$)           & 156 (1.00$\times$)           & 153 (1.02$\times$)           & 138 (1.13$\times$)           & \bfseries  92 (1.70$\times$) \\
                                     & 0.05     & 216 (1.00$\times$)           & 216 (1.00$\times$)           & 216 (1.00$\times$)           & 209 (1.03$\times$)           & \bfseries 129 (1.67$\times$) \\
        \midrule
        \multirow[c]{5}{*}{SVHN}     & 10       & 41 (1.00$\times$)            & 43 (0.95$\times$)            & 41 (1.00$\times$)            & 43 (0.95$\times$)            & \bfseries  40 (1.02$\times$) \\
                                     & 1        & 58 (1.00$\times$)            & 83 (0.70$\times$)            & 61 (0.95$\times$)            & 47 (1.23$\times$)            & \bfseries  44 (1.32$\times$) \\
                                     & 0.5      & 112 (1.00$\times$)           & 185 (0.61$\times$)           & 125 (0.90$\times$)           & 74 (1.51$\times$)            & \bfseries  47 (2.38$\times$) \\
                                     & 0.1      & 265 (1.00$\times$)           & 265 (1.00$\times$)           & 271 (0.98$\times$)           & 320 (0.83$\times$)           & \bfseries 141 (1.88$\times$) \\
                                     & 0.05     & 379 (1.00$\times$)           & 378 (1.00$\times$)           & 379 (1.00$\times$)           & 371 (1.02$\times$)           & \bfseries 191 (1.98$\times$) \\
        \bottomrule
    \end{tabular}
    \caption{Number of global iterations required to reach 90\% of the best accuracy for different FL algorithms on CIFAR-10, MNIST, and SVHN with different levels of heterogeneity.
        The results are averaged over 5 runs.
        The best results for each setting are highlighted in bold.
    }
    \label{tab:epochs_to_80}
\end{table*}

\subsection{Convergence Rate}
Table \ref{tab:epochs_to_80} shows the number of global iterations required to reach 90\% of the best performance of FedAvg in the experiments.
The speed-up factor in brackets is calculated as the ratio of the global iterations required by FedAvg and the selected algorithm.
Except for the easy task MNIST with $\alpha=10$, FedGA achieves the fastest convergence speed in all the cases.
And with larger $\alpha$, FedGA is more robust and converges faster than other FL algorithms.

\begin{figure}[!ht]
    \centering
    \includegraphics[width=0.9\linewidth]{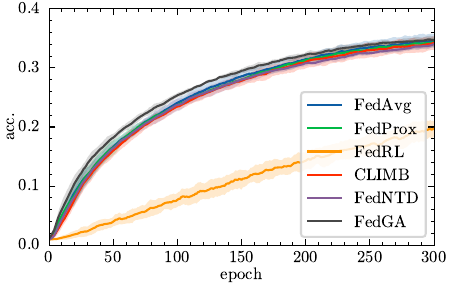}
    \caption{Accuracy on CIFAR-100 with $\alpha=10$ and Adam optimizer.}
    \label{fig:cifar100-10-adam}
\end{figure}

\begin{figure}[!ht]
    \centering
    \includegraphics[width=0.9\linewidth]{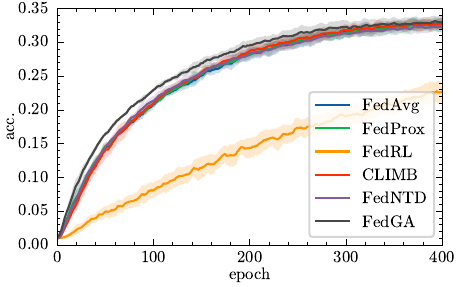}
    \caption{Accuracy on CIFAR-100 with $\alpha=1$ and Adam optimizer.}
    \label{fig:cifar100-1-adam}
\end{figure}

\begin{figure}[!ht]
    \centering
    \includegraphics[width=0.9\linewidth]{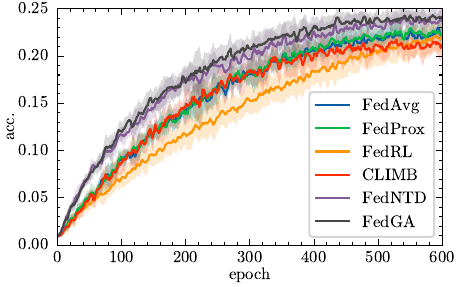}
    \caption{Accuracy on CIFAR-100 with $\alpha=0.1$ and Adam optimizer.}
    \label{fig:cifar100-0.1-adam}
\end{figure}

We also explore the Adam optimizer on CIFAR-100 and Tiny-ImageNet.
The results are shown in Figure~\ref{fig:cifar100-10-adam}, Figure~\ref{fig:cifar100-1-adam}, and Figure~\ref{fig:cifar100-0.1-adam}.
Results on CIFAR-100 datasets shows that GA consistently outperforms or comparable to other FL algorithms in terms of accuracy and F1 score.
With $\alpha \in \{10, 1, 0.1\}$, GA achieves the best performance, while with $\alpha \in \{0.5, 0.05\}$, GA is comparable to FedNTD.


Notably, the performance of FedGA is comparable to FedNTD based on same global iterations.
However, FedNTD requires grid search for two additional hyperparameters, while FedGA is parameter-free and easy to use in practice.
Besides, FedNTD requires an additional global model forward pass to calculate the not-true label, which doubles the computation locally.

Obviously, the accuracy curve of FedGA is constantly higher than other FL algorithms during the training process in all the cases.
With this 100 class classification task, and the Adam optimizer, GA shows its effectiveness in reducing the bias and improving the performance in complex FL scenarios and advanced neural network architectures with adaptive optimizer.

When setting $\alpha=0.1$, performance of FedGA is clearly the upper bound of FedNTD algorithm.

\subsection{Sensitivity to Local Training Epoch}
\label{sec:sensitive_to_local_epoch}

It is well known that the number of local epochs is a critical hyperparameter in FL.
With heterogeneous data distribution, increasing the local epoch may exaggerate the bias of the local model and mislead the server aggregation.
The desired pattern is that more local epochs benefits the training by reducing the global iterations and communication cost.

\begin{figure}[!ht]
    \centering
    \includegraphics[width=.82\linewidth]{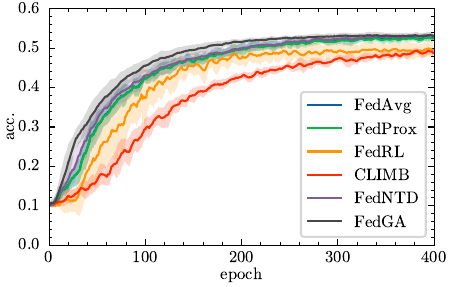}
    \caption{Accuracy on CIFAR-10 with $\alpha=0.5$ and 2 local epochs.}
    \label{fig:cifar10-0.5-2}
\end{figure}

\begin{figure}[!ht]
    \centering
    \includegraphics[width=.82\linewidth]{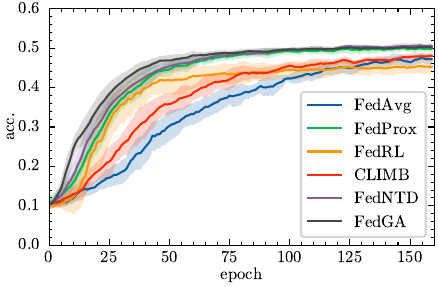}
    \caption{Accuracy on CIFAR-10 with $\alpha=0.5$ and 5 local epochs.}
    \label{fig:cifar10-0.5-5}
\end{figure}

\begin{figure}[!ht]
    \centering
    \includegraphics[width=.82\linewidth]{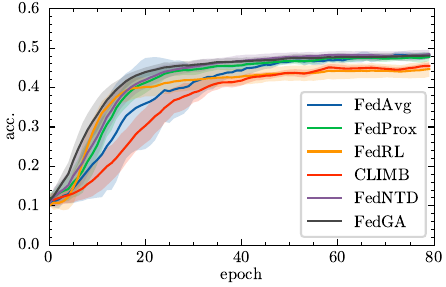}
    \caption{Accuracy on CIFAR-10 with $\alpha=0.5$ and 10 local epochs.}
    \label{fig:cifar10-0.5-10}
\end{figure}

Local epoch in range of $\{2, 5, 10\}$ are tested on CIFAR-10 with $\alpha=0.5$.
The performance of selected FL algorithms are shown in Figure~\ref{fig:cifar10-0.5-2}, Figure~\ref{fig:cifar10-0.5-5}, and Figure~\ref{fig:cifar10-0.5-10}.
Despite the local epoch per global iteration, we keep the total number of local epochs the same for each setting.
Therefore, for local epoch 2, the total local epochs are 400, for local epoch 5, the total local epochs are 160, and for local epoch 10, the total local epochs are 80.
In this case, all experiments are conducted with 800 local epochs in total.

As shown in the tree figures, with more local epochs per global iteration, the performance of all FL algorithms are degraded.
With 2 local epochs, all selected algorithms achieve over 51\% accuracy.
FedGA still converges faster than other FL algorithms in the first half of the training process.
With 5 and 10 local epochs, FedGA achieves the top accuracy and shows convergence advantage constantly.

Generally, the accuracy curve of FedGA is still the highest among all the algorithms.
This indicates that FedGA is less sensitive to the number of local epochs and can maintain the performance with fewer local epochs compared to other FL algorithms.

\bibliography{aaai25}